%% file: iclr2025_conference.tex
\documentclass{article} 
\usepackage{tccml_iclr2025_conference,times}
\pdfoutput=1
\input{math_commands.tex}

\usepackage{hyperref}
\usepackage{url}
\usepackage{graphicx}
\usepackage{float}
\usepackage{siunitx}

\title{Conditional Diffusion-Based Retrieval of \\ Atmospheric CO$_{2}$
           from Earth Observing \\ Spectroscopy }

\author{William R. Keely \\
University of Oklahoma\\
Norman, OK 73019, USA \\
\texttt{william.r.keely@ou.edu} \\
\And
Otto Lamminpää \\
Jet Propulsion Laboratory \\
California Institute of Technology \\
Pasadena, CA 91109, USA \\
\texttt{otto.m.lamminpaa@jpl.nasa.gov} \\
\And
Steffen Mauceri \\
Jet Propulsion Laboratory \\
California Institute of Technology \\
Pasadena, CA 91109, USA \\
\texttt{steffen.mauceri@jpl.nasa.gov}
\And
Sean M. R. Crowell \\
LumenUs Scientific \\
Oklahoma City, OK 73102, USA \\
\texttt{sean@belumenus.com}
\And
Christopher W. O'Dell \\
Cooperative Institute for Atmospheric Research \\
Colorado State University \\
Fort Collins, CO 80521, USA \\
\texttt{christopher.odell@colostate.edu}
\And
Gregory, R. McGarragh \\
Cooperative Institute for Atmospheric Research \\
Colorado State University \\
Fort Collins, CO 80521, USA \\
\texttt{greg.mcgarragh@colostate.edu}
}
\iclrfinalcopy 
\begin{document}

\maketitle

\begin{abstract}
 Satellite-based estimates of greenhouse gas (GHG) properties from observations of reflected solar spectra are integral for understanding and monitoring complex terrestrial systems and their impact on the carbon cycle due to their near global coverage. Known as \emph{retrieval}, making GHG concentration estimations from these observations is a non-linear Bayesian inverse problem, which is operationally solved using a computationally expensive algorithm called Optimal Estimation (OE), providing a Gaussian approximation to a non-Gaussian posterior. This leads to issues in solver algorithm convergence, and to unrealistically confident uncertainty estimates for the retrieved quantities. Upcoming satellite missions will provide orders of magnitude more data than the current constellation of GHG observers. Development of fast and accurate retrieval algorithms with robust uncertainty quantification is critical. Doing so stands to provide substantial climate impact of moving towards the goal of near continuous real-time global monitoring of carbon sources and sinks which is essential for policy making. To achieve this goal, we propose a diffusion-based approach to flexibly retrieve a Gaussian or non-Gaussian posterior, for NASA's Orbiting Carbon Observatory-2 spectrometer, while providing a substantial computational speed-up over the current operational state-of-the-art.
\end{abstract}

\section{Introduction}
Satellite observations of greenhouse gases (GHGs) are essential for monitoring and quantifying the impact of anthropogenic activities (e.g., fossil fuel combustion, deforestation, urbanization) on the climate of the planet particularly in rapidly changing regions like the high latitudes and Tropics which are difficult to measure \emph{in situ}. Satellite-bourne spectrometers such as NASA's Orbiting Carbon Observatory-2 and 3 (OCO-2/3; \citet{crisp2015measuring}), and JAXA's Greenhouse Gases Observing Satellite (GOSAT; \citet{yokota2009global}) make high spectral resolution measurements of reflected solar spectra in order to infer changes in GHGs. A physics-based algorithm is used to estimate or \textit{retrieve} the atmospheric state including the column averaged dry air-mole fraction of atmospheric carbon dioxide (XCO$_{2}$) from these spectra. XCO$_{2}$ observations are used to constrain surface fluxes of GHGs, which are further used to study the Earth's complex carbon cycle. \citep{crowell20192015}.

Estimating GHG concentrations from spectral observations is typically posed as an inverse problem, where we aim to recover the state of the atmosphere, denoted $\mathbf{x}$, from observed spectra $\mathbf{y}$ with instrument viewing geometries $\mathbf{b}$. The relationship between state and observation can be written as: $\mathbf{y} = F(\mathbf{x, b}) + \mathbf{\nu}$, where $F$ is the forward operator, and noise $\mathbf{\nu} \sim \displaystyle \mathcal{N}(0, \mathbf{R})$ which is scene and wavelength-dependent. OCO-2 uses an Optimal Estimation (OE) based algorithm called Atmospheric Carbon Observations from Space (ACOS) detailed in \citet{connor2008orbiting}; \citep{odell2012acos}. OE methods solve the inverse problem using Bayes Rule: $p(\mathbf{x}|\mathbf{y})=p(\mathbf{y}|\mathbf{x})p(\mathbf{x})$, where estimates of $\mathbf{x}$ are obtained as a \textit{maximum a posteriori} (MAP) point estimate which maximizes the posterior probability $p(\mathbf{x}|\mathbf{y})$ \citep{rodgers2000inverse} with an associated Gaussian approximation to the posterior (co-)variance as an uncertainty estimate. The ACOS retrieval employs a Radiative Transfer Model forward model (FM) coupled with an instrument model for the OCO-2 spectrometer as $F$ to produce simulated spectra, $\mathbf{y_{Sim}}$, and partial derivatives for a given $\mathbf{x}$. The algorithm iteratively seeks the MAP estimate that minimizes a cost function (Appendix \ref{AppendixOE}) that represents the sum of squared errors between the observations $\mathbf{y_{Obs}}$ and $\mathbf{y_{Sim}}$.

\citet{hobbs2017simulation}, \citet{lamminpaa2019mcmc} and \citet{braverman2021post} illustrate how realizations of the true conditional posterior for XCO$_{2}$ can be non-Gaussian or bimodal due to the nonlinear nature of the forward model, arising from varying aerosol and surface conditions. The classical formulation does not apply in these cases. The Gaussian assumed uncertainty estimate produced by ACOS is often over confident, e.g., the prediction interval often does not contain the true value of XCO$_{2}$. Applications of these data to constrain CO$_2$ emissions can be sensitive to the assumed uncertainties and often incorporate ad hoc adjustments to the uncertainties prior to their use. This introduces additional uncertainties in the CO$_2$ emissions estimates themselves \citep{peiro2022four}. The uncertainty in predicted emissions is critical for inference at policy relevant (e.g., national) scales necessary for treaties and enforcement \citep{byrne2022national}.

OE requires repeated evaluations of the computationally expensive forward FM per observation. OCO-2 currently makes $10^{5}$ cloud-free observations per day and has a record that spans from 2014 to present. The decade-long catalog has undergone several updates, with reprocessing taking around a year to complete with the current ACOS version and modern HPC architectures. This computational bottleneck will be prohibitive for future missions such as ESA's upcoming Carbon Dioxide Monitoring Mission \citep{sierk2021copernicus}, which will generate tens of millions of observations daily.  

It is critical to develop new retrieval techniques that are fast and accurate while also providing robust uncertainties. In this work, we apply a diffusion-based approach for accurate atmospheric CO$_2$ retrieval that avoids the need for repeated FM evaluations after training is complete, while also providing a robust uncertainty for NASA's OCO-2 instrument. Diffusion is a generative machine learning method that allows for efficient sampling of potentially complex posteriors and has seen application in solving inverse problems in the natural sciences. Diffusion's success comes from the ability to define a prior distribution which is typically a standard Gaussian $\displaystyle \mathcal{N}(0,\displaystyle \mI)$ and then training a \emph{denoising} model that iteratively learns to shift samples drawn from the prior to the true data distribution using a Markov process. This allows for efficient sampling to recover either a Gaussian uncertainty by taking the first two statistical moments of the draws or to explore non-Gaussianity in the retrieved posterior.

\section{Methods}

To train the retrieval we use the L2 product \citep{payne2023dug} of version 11 ACOS FM simulated spectra denoted as $\mathbf{y_{Sim}}$. OCO-2's wavelength ranges are the near-infrared Oxygen A band near \SI{0.76}{\micro\metre}, the shortwave infrared 'weak' CO$_{2}$ band near \SI{1.6}{\micro\metre}, and the shortwave infrared 'strong' CO$_{2}$ band near \SI{2.05}{\micro\metre}. Additional retrieval covariates include the solar zenith angle ($SZA$) and a weather model predicted surface pressure term $P_{Surf}$ that comes from analysis produced by the GEOS5-FP-IT data assimilation system \citep{lucchesi2013file}.  The concatenated covariate vector for training point $i$ is denoted as $\mathbf{\bar{y_{i}}} = \{ \mathbf{y_{i}}, SZA_{i}, P_{Surf_{i}}\}$. The XCO$_{2}$ scalars from the L2 state vector $\mathbf{x}$, in parts per million (ppm), serve as the training label $x_{i}$. In total, we construct a training dataset of $3 \times 10^6$ $\{ \mathbf{\bar{y_{i}}}, x_{i}\}$ pairs. Dependence on the FM for generating training data can introduce errors into the retrieval due to differences between simulated training radiance and real observations seen at inference time: ($\mathbf{y_{Sim}}$ $-$ $\mathbf{y_{Obs}}$). We account for this difference by careful perturbation of the simulated radiance prior to model training, as shown in Appendix \ref{AppendixEOF} and resample both $\mathbf{y_{Sim}}$ and $\mathbf{y_{Obs}}$ to a common wavelength grid. We use the FM for training despite the abundance of OCO-2 observations with an eye to current and future missions that may lack sufficient temporal or spatial coverage of observations for an ML-based retrieval. These missions cannot adopt ML algorithms trained directly on observations to generalize well across the globe, thus requiring learning on simulated training data.

Given realizations of the covariates, we aim to retrieve the posterior distribution $p(x | \mathbf{\bar{y}})$, using the conditional diffusion method described in \citet{han2022card}. This paper suggests the use of a conditional mean prior, $x_{prior}$, which is a pre-trained neural network $f_{\phi}$ with weights $\phi$ to parameterize the forward diffusion distribution as $\displaystyle \mathcal{N}(x_{prior}, 1)$ instead of the more standard choice of $\displaystyle \mathcal{N}(0, 1)$. This introduces conditioning information into the forward process and potentially speeds up inference by reducing the number of diffusion steps required. We explore different methods to specify $x_{prior}$ including a neural network retrieval of XCO$_{2}$ as well as XCO$_{2}$ values from GEOS-FP-IT in Appendix \ref{AppendixCondMean}. The denoising model, $\epsilon_{\theta}(\mathbf{\bar{y_{i}}}, x_{prior_{i}}, x_{i,t}, t)$ is then trained as described in \citet{ho2020denoising} using the mean squared error loss, where $x_{t}$ is the noised label at diffusion step $t$. 

The OCO-2 operational XCO$_2$ product includes an extensive post hoc bias correction to reduce systematic errors or \textit{biases} in the raw OE retrieved XCO$_{2}$ \citep{keely2023nonlinear,odell2018improved}. This process involves fitting a regression model on errors in XCO$_{2}$ with predictors selected from the retrieved atmospheric state. XCO$_2$ errors are calculated using ground truth labels of XCO$_{2}$ from the Total Carbon Column Observing Network (TCCON; \citep{wunch2011total}) that are collocated with OCO-2 measurements in space and time. In this spirit, we evaluate the potential of reducing bias in the trained diffusion model using a similar dataset of collocated TCCON XCO$_{2}$ labels to finetune the retrieved value. We collocate OCO-2 and TCCON samples within 100km spatially and 1 hour temporally, for which we also have FM produced radiance for years 2015 to 2021. This results in a finetuning dataset of $25 \times 10^4$ pairs. To finetune the retrieval, we first unfreeze the last layers of $f_{\phi}$ and train an additional 10 epochs. We then continue training $\epsilon_{\theta}$ using the finetuned conditional mean backbone until the loss for a small validation set stabilizes. 

To evaluate the performance of the proposed diffusion retrieval relative to ACOS, we use a holdout set of data of OCO-2 observations collocated with TCCON sites and retrieve on the OCO-2 measured reflected solar radiance ($\mathbf{y_{Obs}}$) for the year of 2022. This test dataset contains 20,000 collocated pairs, and contains examples from all operational TCCON sites for the year. We assess the quality of the point estimates of XCO$_{2}$ for ACOS and the diffusion retrieval with and without finetuning/bias-correction, as well as the quality of their uncertainty estimates.

\section{Results}

\subsection{Efficient Posterior Sampling:}
The reverse diffusion process allows us to generate samples from $p(x|\mathbf{\bar{y}})$ by first drawing each $x_{T}\sim{\displaystyle \mathcal{N}(x_{prior},1)}$ and then iteratively denoising with $\epsilon_{\theta}$ from $t=T$ to $t=0$ to recover the denoised XCO$_{2}$ samples, thus enabling efficient density estimation of potentially complex posteriors as shown in Figure 1. The diffusion retrieval takes on average 0.0003 seconds to draw 10
conditional samples for a single observation on a 2017 Apple MacBook Pro. \citet{lamminpaa2024forward} showed that FM evaluations on similar hardware take approximately 33-55 seconds. 

\begin{figure}[h]
\begin{center}
\includegraphics[width=0.5\linewidth]{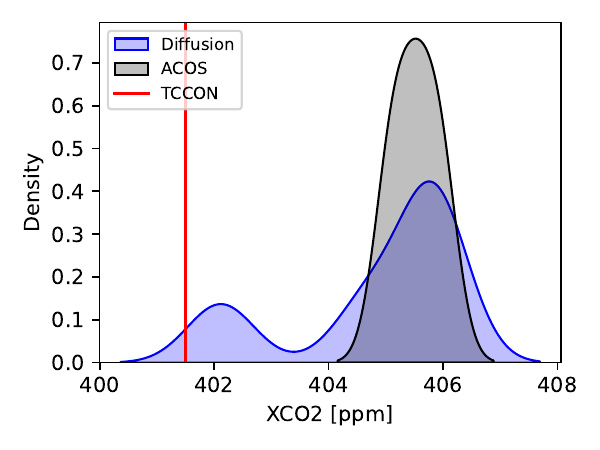}
\label{posterior}
\end{center}
\caption{Example of bi-modal distribution recovered through 10 conditional samples from the Diffusion posterior (blue) without finetuning compared to the operational OE posterior which is derived from the point estimate $\pm$ an estimate of the variance  (black) without bias correction. The Diffusion posterior covers the ground truth value from TCCON (red) within the first mode which is not captured by the Gaussian operational posterior.}
\end{figure}

\subsection{Comparison to the Operational OCO-2 Retrieval:}
In Figure 2, we evaluate the XCO$_{2}$ point estimate and uncertainty for both the operational ACOS retrieval and our proposed diffusion-based retrieval on a holdout year of OCO-2 observations collocated with TCCON sites for 2022. The diffusion-based retrieval without finetuning shows a 3\% improvement to ACOS prior to bias correction. When we compare to the operational product with additional finetuning for the diffusion retrieval, the finetuned diffusion retrieval shows a 10\% improvement in RMSE over bias-corrected ACOS retrieved XCO$_{2}$ for the test year. We also compare the uncertainty estimate of both methods. Since the ACOS retrieval enforces a Gaussian form, we also assume a Gaussian prediction interval for the diffusion retrieval by taking the first two moments over the diffusion samples. The performance of the uncertainties is assessed by calculating the Miscalibration Area which is the integrated region between the calibration curve and the $x=y$ line. The smaller the area, the better the coverage of the prediction interval for a given percentile. The assumed Gaussian prediction interval from the diffusion retrieval contains the collocated TCCON XCO$_{2}$ value more frequently than the operational retrieval. Finetuning provides an added benefit of further calibrating the uncertainty and reducing the Miscalibration Area on the holdout year. In \ref{AppendixMLeval}, we train several additional types of deep learning models that can provide uncertainty quantification of their estimate and compare them to diffusion and ACOS. 

\begin{figure}[h]
\begin{center}
\includegraphics[width=\linewidth]{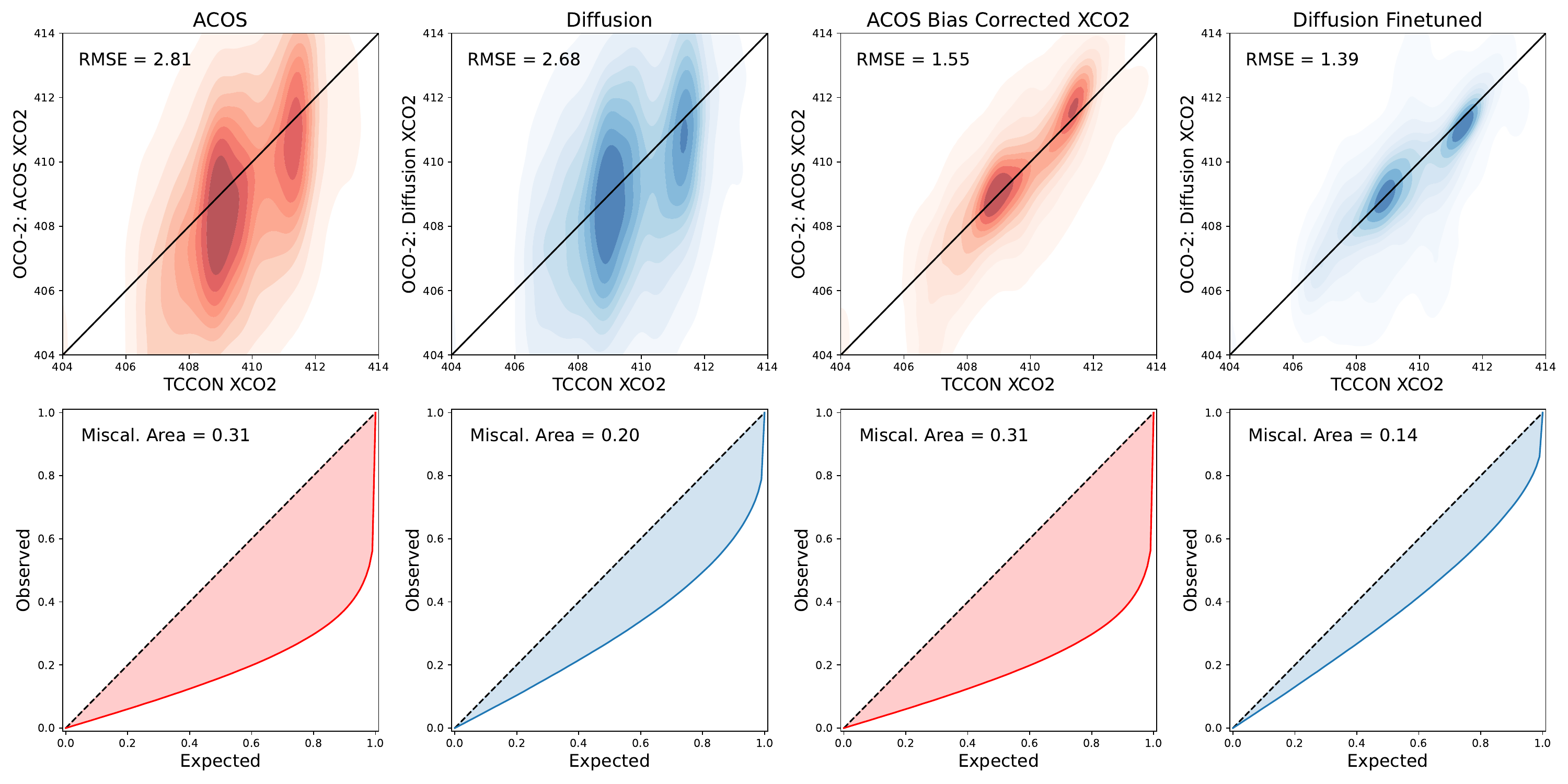}
\label{finetune}
\end{center}
\caption{Comparison of ACOS and Diffusion with and without bias correction/finetuning for the holdout year of 2022. We compare the point estimate of each method (top row) by evaluating the RMSE. We compare the Gaussian uncertainties of each method (bottom row) by evaluating the Miscalibration Area.}
\end{figure}

\section{Conclusions and Future Work}
We present a diffusion-based approach for retrieving atmospheric XCO$_{2}$ from satellite observations that is competitive with the current state-of-the-art approach while offering the for an increase of several orders of magnitude in speed during inference. The ability to quickly sample the conditional posterior also allows for improved uncertainty quantification compared to the operational product, thus allowing for an improved constraint of flux estimates by inversion models. In future work, we will further benchmark the computational speed up against the operational retrieval as well as derive an averaging kernel which is essential for assimilation of XCO$_{2}$ observations into carbon flux inversion models. It will also be important to assess the potential for the finetuning process to introduce biases in regions where TCCON is not available, since regionally coherent systematic errors in XCO$_2$ propagate directly into inferred CO$_2$ surface fluxes. The approach is highly generalizable to retrieval of other important GHGs such as methane (CH$_{4}$) and to future GHG missions such as Carbon-I \citep{frankenbergcarbon} and CO2M, which will produce large data sets directly impacting future climate policy.

\subsubsection*{Acknowledgments}
We would like to deeply thank the OCO-2/3 Science and Algorithm Teams.

We appreciate the TCCON teams, who measure and provide ground-based XCO$_{2}$ validation to the carbon cycle research community (\citep{de2014tccon}; \citep{te2022tccon}, \citep{liu2023tccon}; \citep{hase2022tccon}; \citep{kivi2014tccon}; \citep{weidmanntccon}; \citep{zhou2022tccon}; \citep{deutschertccon}; \citep{dubey2022tccon}; \citep{wunch2022tccon}; \citep{garcia2022tccon}; \citep{shiomi2022tccon}; \citep{strong2022tccon}; \citep{notholt2022tccon}; \citep{morino2022tccon_a}; \citep{morino2022tccon_b}; \citep{morino2022tccon_c}; \citep{sussmann2023tccon}; \citep{buschmann2022tccon}; \citep{sherlock2022tccon_a}; \citep{wennberg2022tccon_a}; \citep{wennberg2022tccon_b})

This work was funded by NASA grant: \textbf{80NSSC24K0763}.

\subsection*{Data Availability}
OCO-2 data including the forward model spectra are available at: \url{https://disc.gsfc.nasa.gov/datasets/OCO2_L2_Diagnostic_11r/summary?keywords=oco2%20diagnostic}. TCCON data is available at: \url{https://tccon-wiki.caltech.edu/Main/DataDescriptionGGG2020}.
We also provide the predictions and Gaussian uncertainty from the diffusion model with collocated TCCON and ACOS estimates here: \url{10.5281/zenodo.15115961}

\bibliography{iclr2025_conference}
\bibliographystyle{iclr2025_conference}

\appendix
\section{Related Work}
\label{AppendixRelated}
Machine learning approaches to greenhouse gas retrieval from Earth observing spectroscopy have made positive impact in reducing the computational cost over full physics based approaches. For OCO-2, \citet{David2021ml} and \citet{breon2022potential} offered a first attempt at directly retrieving column CO$_{2}$ by pairing XCO$_{2}$ labels from the carbon inversion model CAMS directly with observations. Their approach improved computational speed but did not improve on precision due to the coarse scale of the CAMS labels. \citet{keely2022towards} introduced an initial attempt at providing a robust uncertainty quantification for a deep learning retrieval through ensembling but on a limited simulated training data set. \citet{xie2024fast} illustrated the limitations of training an ML retrieval for OCO-2 using observations only, and enhanced the training data set using a fast FM emulator.  \citet{lamminpaa2024forward} proposes emulating the FM with a Gaussian Process to speed-up the iterative forward model evaluations in OE. The non-Gaussianity of posterior distribution of OCO retrieval has been explored using MCMC (\citet{brynjarsdottir2018mcmc}, \citet{lamminpaa2019mcmc}) and Simulation Based UQ (\citet{turmon2019}, \cite{braverman2021post}). In this work, we propose a diffusion-based approach to efficiently sample the conditional posterior and train directly on radiance produced by the high fidelity FM used by the current operational retrieval algorithm for OCO-2. We also show how finetuning the retrieval using co-sampled ground truth labels from TCCON \citep{wunch2011total} can make ML retrievals competitive with the current state-of-the-art. 

\section{Experiments}
\subsection{Empirical Orthogonal Functions}
\label{AppendixEOF}
A matrix of training residuals $M_{j}$ is given for each band $j$. $M_{j}$ is decomposed into its eigenvectors using the singular value decomposition $M_{j}=U_{j}W_{j}V_{j}^{T}$. The $k=1$ eigen vector explains the largest fraction of the total variance. For the OCO-2 operational retrieval, coefficients of these fixed EOFs are retrieved to minimize the effect of errors from forward model discrepancies during retrieval. Hence, they serve as a proxy for the forward model error in the simulated radiances. To reduce similar errors in the ML retrieval we retrain using the EOFs to perturb the simulated training set ideally making the model more robust to sim-to-real differences. Before application to each $i^{th}$ training point, each EOF $\mathbf{e_{j,k}}$ is randomly scaled by a coefficient $c_{i,j,k}$ drawn from the distributions formed by the operational scaling terms. Thus, the training radiance for band $\mathbf{y_{Sim}}$ become:

\begin{equation}
    \mathbf{y_{Sim_{i,j}}^{'}}=\mathbf{y_{Sim_{i,j}}}+\sum_{k=1}^{K}c_{i,j,k}\mathbf{e_{j,k}}
\end{equation}

Figure 3 shows the mean difference between the simulated radiance and observation pairs withheld from fitting of the EOFs. ACOS includes the coefficients of the EOFs in the OE state vector. In future work we will look at retrieving the scaling term in a similar fashion using the diffusion-based retrieval as part of an expanded retrieved state. In Table 1, we perform a simple ablation to quantify the effects of including the Sim-to-Real mitigation during training of an MLP retrieval. By multiplying the EOFs by a random scaler drawn from the scaling distributions of those from the OE retrieval we see a 8\% reduction in error versus the 2022 TCCON evaluation set.

\begin{figure}[h]
\begin{center}
\includegraphics[width=0.9\linewidth]{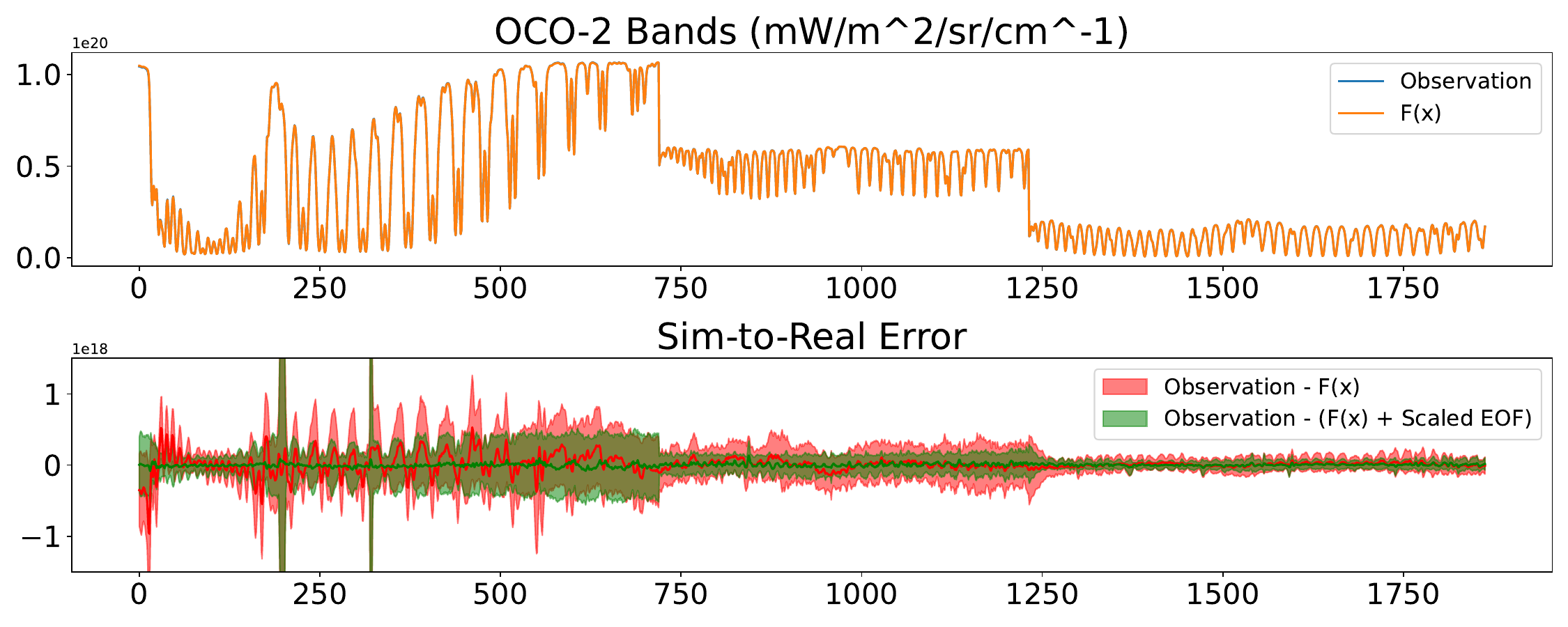}
\label{sim2real}
\end{center}
\caption{The top figure shows an example of an observed radiance (blue) and the FM modeled radiance (orange) for a single sounding. The bottom figure shows the mean and variance of remaining error before and after application of the randomly scaled EOFs.}
\end{figure}

\begin{table}[h]
\caption{Ablation of retrieval with EOF perturbed training radiances and without. RMSE in parts per million [ppm] is calculated for the 2022 TCCON set.}
\label{raw-result}
\begin{center}
\begin{tabular}{ll}
\multicolumn{1}{c}{\bf Retrieval} &\multicolumn{1}{c}{\bf RMSE [ppm]}
\\ \hline \\
MLP + EOF & 3.15 \\
MLP + Randomly Scaled EOF & \textbf{3.11} \\
MLP Baseline & 3.37 \\
\end{tabular}
\end{center}
\end{table}

\subsection{The Conditional Mean Prior}
\label{AppendixCondMean}
In Table 2, we evaluate several choices for the conditional mean prior, $x_{prior}$, in the forward and reverse diffusion processes. First, we follow the method outlined in \citet{han2022card}, where $x_{prior} = f(\mathbf{y})_{\phi}$ and $f_{\phi}$ is a pre-trained MLP model. For the second approach we modify the MLP to use 1-dimensional convolution layers with max-pooling. Lastly, we evaluate using the GEOS-FP-IT \citep{lucchesi2013file} XCO$_{2}$ labels which are also used as priors in the operational OE-based retrieval. We find that the convolutional model offers the best performance, where the other two priors provide a poorer RMSE as compared to the ACOS retrieval (shown in Table 3).

\begin{table}[h]
\caption{Comparison of diffusion retrievals with different conditional mean priors. Mean and Standard Deviation of bias in parts per million [ppm] is calculated for the 2022 TCCON set.}
\label{finetuned-results}
\begin{center}
\begin{tabular}{ll}
\multicolumn{1}{c}{\bf Cond. Mean Prior}  &\multicolumn{1}{c}{\bf Bias [ppm]}
\\ \hline \\
MLP & -0.70 $\pm$ 2.88  \\
Conv1D & \textbf{-0.23 $\pm$ 2.77 } \\
GEOS-FP-IT & -1.03 $\pm$ 2.95 \\
\end{tabular}
\end{center}
\end{table}

\subsection{Comparison of Diffusion with alternative ML retrievals}
\label{AppendixMLeval}
In Table 3, we compare the conditional diffusion retrieval to several other deep learning-based methods that provide an uncertainty of their point estimates. The choice of methods are state of the art in several different paradigms of UQ for deep learning models: ensemble, kernel learning, and hierarchical Bayesian.

\begin{table}[h]
\caption{Mean and Standard Deviation of bias in parts per million [ppm] and Miscalibration Area (Miscal. Area) for ML retrieval calculated for the 2022 TCCON set.}
\label{raw-result}
\begin{center}
\begin{tabular}{lll}
\multicolumn{1}{c}{\bf Method}  &\multicolumn{1}{c}{\bf Bias [ppm]} & \multicolumn{1}{c}{\bf Miscal. Area}
\\ \hline \\
ACOS    & -0.62 $\pm$ 2.82 & 0.31 \\
MLP     & -0.8 $\pm$ 3.33 & N/A \\
SWAG    & -0.56 $\pm$ 3.14 & 0.23 \\
Deep Kernel & -1.54 $\pm$ 4.50 & 0.38 \\
Evidential    & -0.49 $\pm$ 2.80 & 0.26 \\
Diffusion     & \textbf{-0.23 $\pm$ 2.72} & \textbf{0.21} \\
\end{tabular}
\end{center}
\end{table}

\textbf{SWAG:} Stochastic Weighted Averaging-Gaussian or SWAG is a Bayesian extension of Deep Ensembles \citep{lakshminarayanan2017simple} by approximating a Bayesian Model Average through a learned distribution over multiple Stochastic Gradient Descent iterates \citep{maddox2019swag}.

\textbf{Evidential Regression:} Deep Evidential Regression \citep{amini2020deep} captures both aleatoric and epistemic sources of uncertainty with a single model by placing hyper-priors over the Gaussian likelihood and employing a Inverse-Gamma distribution over the prior to form a Normal Inverse-Gamma conjugate.

\textbf{Deep Kernel Learning:} Deep Kernel Learning \citep{wilson2016deep} offers a highly scalable approach to Gaussian Processes by parameterizing a kernel such as a Matern with a neural network and provides an epistemic uncertainty of its point estimates.

The diffusion model is implemented using MLP architectures with \textit{Conv1D} layers for both $f_{\phi}$ and $\epsilon_{\theta}$. We use a $T=50$ with a cosine noise schedule \citep{nichol2021improved}. The remaining parameters for the start and end values for the schedule and EMA parameters are left as suggested in \citet{han2022card}. The retrieved XCO$_{2}$ from the diffusion model is closer to a mean bias of 0.0 parts-per-million than the operational ACOS retrieval without any finetuning or bias correction with a similar error variance. We also note that both the Evidential and SWAG retrievals provide an improved comparison to the TCCON ground truth than ACOS. These methods do not require a conditional mean prior, as observed in Table 2. which can have a substantial impact on the diffusion-based retrieval.

\section{The Optimal Estimation Cost Function}
\label{AppendixOE}
The ACOS retrieval employs a Radiative Transfer Model forward model (FM) coupled with an instrument model for the OCO-2 spectrometer as $F$ to produce simulated spectra and partial derivatives for a given $\mathbf{x}$. The algorithm iteratively seeks the MAP estimate that minimizes the cost function that represents the sum of squared errors with the observations $\mathbf{y}$ together with a prior state vector $\mathbf{x}_a$, both scaled by their uncertainties expressed as covariance matrices:
\begin{equation}
    J(\mathbf{x})=(\mathbf{y}-F(\mathbf{x},\mathbf{b}))^{\mathsf{T}}\mathbf{R}^{-1}(\mathbf{y}-F(\mathbf{x},\mathbf{b})) + (\mathbf{x}-\mathbf{x}_a)^{\mathsf{T}}\mathbf{B}^{-1}(\mathbf{x}-\mathbf{x}_a), 
\end{equation}
where $\mathbf{R}$ is the covariance matrix of the noise $\nu \sim \displaystyle \mathcal{N}(0, \mathbf{R})$, the prior is Gaussian with mean $\mathbf{x}_a$ and covariance $\mathbf{B}$, i.e. $\mathbf{x} \sim \displaystyle \mathcal{N}(\mathbf{x}_a, \mathbf{B})$. The minimizer $\mathbf{\hat{x}}$ of $J$ is the MAP estimate $\arg\max_{\mathbf{x}}\exp(-\frac{1}{2}J(\mathbf{x}))$, and under the assumption of Gaussianity the posterior uncertainty has an analytic form, namely, $\left(\left(\frac{d}{dx}F(\mathbf{\hat{x}},\mathbf{b})\right)^T \mathbf{R}^{-1}\left(\frac{d}{dx}F(\mathbf{\hat{x},\mathbf{b}})\right) + \mathbf{B}^{-1}\right)^{-1}$ \citep{rodgers2000inverse}.

\end{document}

%% file: math_commands.tex
\usepackage{amsmath,amsfonts,bm}

\def\eqref#1{equation~\ref{#1}}

\def\1{\bm{1}}

\def\mI{{\bm{I}}}

\DeclareMathAlphabet{\mathsfit}{\encodingdefault}{\sfdefault}{m}{sl}
\SetMathAlphabet{\mathsfit}{bold}{\encodingdefault}{\sfdefault}{bx}{n}

